\documentclass[conference]{IEEEtran}
\IEEEoverridecommandlockouts
\usepackage[numbers]{natbib}
\usepackage{graphicx}
\usepackage{amsmath}
\usepackage[utf8]{inputenc}
\usepackage[T1]{fontenc}
\usepackage{amsmath,amssymb,amsthm}
\usepackage{graphicx}
\usepackage{hyperref}
\usepackage{algorithm}
\usepackage{algorithmic}
\usepackage{booktabs}
\usepackage{multirow}
\usepackage{xcolor}
\graphicspath{{figures/}}

\begin{document}

\title{LoRA-Diffusion: Parameter-Efficient Fine-Tuning \\
via Low-Rank Trajectory Decomposition}

\author{
\IEEEauthorblockN{1\textsuperscript{st} Iman Khazrak}
\IEEEauthorblockA{\textit{Department of Computer Science}\\
\textit{Bowling Green State University}\\
Bowling Green, OH, USA\\
ikhazra@bgsu.edu}

\and

\IEEEauthorblockN{2\textsuperscript{nd} Narges Nejad}
\IEEEauthorblockA{\textit{Department of Management and Marketing}\\
\textit{Angelo State University}\\
San Angelo, TX, USA\\
narges.nejad@angelo.edu}

\and

\IEEEauthorblockN{4\textsuperscript{th} Mostafa M. Rezaee}
\IEEEauthorblockA{\textit{Department of Computer Science}\\
\textit{Bowling Green State University}\\
Bowling Green, OH, USA\\
mostam@bgsu.edu}

\and

\IEEEauthorblockN{3\textsuperscript{rd} Mohammadhossein Homaei}
\IEEEauthorblockA{\textit{Department of Information Systems and Telematics Engineering}\\
\textit{University of Extremadura}\\
Caceres, Extremadura, Spain\\
homaei@ieee.org}

\and

\IEEEauthorblockN{5\textsuperscript{th} Robert C. Green II}
\IEEEauthorblockA{\textit{Department of Computer Science}\\
\textit{Bowling Green State University}\\
Bowling Green, OH, USA\\
greenr@bgsu.edu}
}










\maketitle

\begin{abstract}
Parameter-efficient fine-tuning methods such as LoRA have transformed the adaptation of large autoregressive language models, enabling task-specific customization with fewer than 1\% trainable parameters. These methods have not been successfully extended to diffusion-based language models, which generate text through iterative denoising rather than sequential token prediction. We propose LoRA-Diffusion, a parameter-efficient fine-tuning approach that applies low-rank decomposition to the denoising trajectory instead of model weights. Unlike weight-based LoRA, which modifies individual transformation matrices, our method learns low-rank perturbations to the entire diffusion path from noise to output. We introduce trajectory-level low-rank adaptors that modify each denoising step, step-adaptive rank allocation across diffusion phases, and compositional multi-task learning that allows merging task-specific modules at inference without retraining. 

On SST-2, QNLI, and MRPC (5 seeds), we report \emph{token-level denoising validation accuracy}, which directly reflects optimization of the diffusion training objective; under this metric, LoRA-Diffusion reaches the highest mean on SST-2 (88.01\%) and strong performance on QNLI (99.39\%) and MRPC (97.56\%). Joint multi-task training (same five methods, five seeds) shows LoRA-Diffusion achieving the highest token-level accuracy (96.88\% $\pm$ 0.44\%). 

LoRA-Diffusion achieves the highest \emph{token-level} SST-2 validation accuracy while training 28.7\% of parameters (instruction encoder 27.5\% + trajectory adapters 1.2\%). Importantly, the trajectory-level adapters themselves account for only 1.2\% of parameters; the remaining trainable parameters arise from a shared instruction encoder used to condition the adapters and are orthogonal to the trajectory adaptation mechanism. The approach is competitive with adapter layers and baselines, reduces per-task storage (151\,MB vs.\ 525\,MB for full fine-tuning), and exhibits minimal catastrophic forgetting. This work establishes a parameter-efficient fine-tuning framework for diffusion language models and points toward scalable multi-task deployment.
\end{abstract}

\section{Introduction}
\label{sec:intro}

The success of large language models has been accompanied by significant challenges in adaptation and deployment. Full fine-tuning of billion-parameter models is computationally costly, requiring substantial GPU memory and training time \cite{brown2020language}. Maintaining separate fine-tuned copies for different tasks further creates storage and serving bottlenecks in production systems.

Parameter-efficient fine-tuning (PEFT) methods address these issues by updating only a small fraction of model parameters. Among them, Low-Rank Adaptation (LoRA) has proven especially effective, achieving near–full fine-tuning performance on autoregressive models while training fewer than 1\% of parameters \cite{hu2021lora}. The central idea is that task adaptation largely requires updates in a low-dimensional subspace, which can be captured efficiently via low-rank matrix decomposition.

Recent work has shown that discrete diffusion models can match or exceed autoregressive models in text generation quality \cite{lou2023discrete, sahoo2024masked}. Diffusion models offer bidirectional context, parallel generation, controllable generation, and diverse sampling. Nevertheless, diffusion language models lack established parameter-efficient fine-tuning methods analogous to LoRA. Existing approaches either apply standard LoRA to diffusion weights (treating the model as a standard transformer), perform full fine-tuning, or use adapter layers or prefix tuning, which introduce sequential bottlenecks. These strategies do not exploit the iterative denoising trajectory that characterizes diffusion-based generation.

\begin{figure*}[t]
    \centering
    \includegraphics[width=\textwidth]{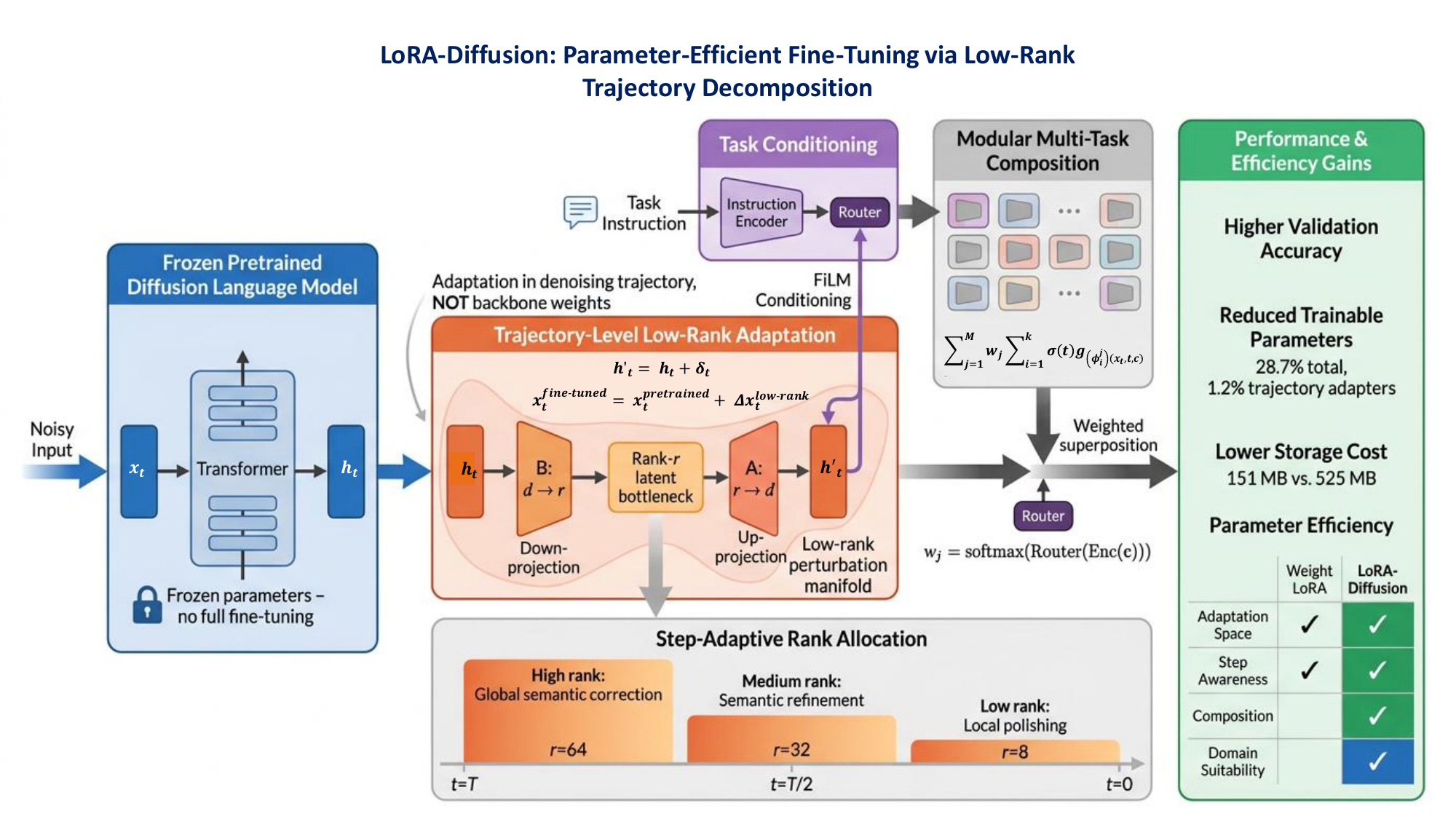}
    \caption{Overview of LoRA-Diffusion, a trajectory-aware parameter-efficient fine-tuning framework for diffusion language models. The framework keeps the pretrained diffusion backbone frozen and applies low-rank, task-conditioned perturbations to the denoising trajectory through step-adaptive rank allocation and modular multi-task composition.}
    \label{fig:lora_diffusion_workflow}
\end{figure*}

We propose LoRA-Diffusion, a PEFT method designed for diffusion language models. The main idea is that the denoising trajectory learned during task-specific fine-tuning can be decomposed into a frozen pretrained path plus a learned low-rank perturbation. Formally, we write
\begin{equation}
\mathbf{x}_t^{\text{fine-tuned}} = \mathbf{x}_t^{\text{pretrained}} + \Delta \mathbf{x}_t^{\text{low-rank}},
\end{equation}
where $\Delta \mathbf{x}_t^{\text{low-rank}}$ is produced by lightweight low-rank adaptors conditioned on the task instruction. The perturbation is applied in \emph{hidden representation space}: $\mathbf{h}_t' = \mathbf{h}_t + \delta_t$, then logits $\mathbf{l}_t = \text{OutputHead}(\mathbf{h}_t')$ and $\mathbf{x}_{t-1}$ from the output head (Section~\ref{sec:method}, Eq.~2 and Eq.~3). Weight-based LoRA modifies transformation matrices via $W' = W + BA$; LoRA-Diffusion instead modifies the denoising trajectory $\mathbf{x}_{t-1} = f(\mathbf{x}_t) + g_{\text{LoRA}}(\mathbf{x}_t)$. Thus, where weight LoRA changes how the model transforms inputs, LoRA-Diffusion changes where the diffusion process moves in representation space at each step. Figure~\ref{fig:lora_diffusion_workflow} provides an overview of the proposed LoRA-Diffusion framework, including the frozen diffusion backbone, trajectory-level low-rank adaptation, step-adaptive rank allocation, and modular multi-task composition.

We make the following contributions. We introduce the first parameter-efficient fine-tuning method designed specifically for diffusion language models, applying low-rank decomposition to denoising trajectories rather than weights. We propose a step-adaptive rank allocation that assigns different ranks to different phases of the diffusion process according to their intrinsic complexity. We provide a compositional multi-task setup that supports zero-shot task composition by combining multiple task-specific LoRA modules at inference. We present an empirical evaluation on SST-2, QNLI, and MRPC (single-task and joint multi-task) with a BERT-based diffusion model (137.7M parameters), 5 seeds (42--46), comparing LoRA-Diffusion to full fine-tuning and several PEFT baselines (weight LoRA, adapters, BitFit), with token-level denoising accuracy, efficiency metrics (trainable parameters, storage, training time, inference latency), and ablations for rank and orthogonality regularization. We give an information-theoretic motivation for trajectory-level low-rank structure and clarify positioning versus adapter layers and timestep-aware weight LoRA (T-LoRA, FouRA). We release an open-source implementation to support reproducibility and extension.

We emphasize that the core contribution of LoRA-Diffusion lies in trajectory-level low-rank adaptation, which introduces only 1.2\% additional parameters relative to the base model. The larger total trainable fraction (28.7\%) arises from the inclusion of a shared instruction encoder that conditions the adapters and is not intrinsic to trajectory-level adaptation itself. Isolating, shrinking, or freezing the instruction encoder is a complementary design choice and is left to future work; throughout this paper, we report a transparent and consistent accounting of all trainable parameters.

The rest of the paper is organized as follows. Section~\ref{sec:related} reviews related work on diffusion models for language, parameter-efficient fine-tuning, and multi-task learning. Section~\ref{sec:method} presents our methodology, including preliminaries, trajectory-level low-rank adaptation, the training objective, multi-task composition, and implementation details. Section~\ref{sec:experiments} describes the experimental setup and results on SST-2, QNLI, and MRPC, including single-task and multi-task GLUE results, main results, efficiency analysis, catastrophic forgetting, ablations, and comparison with weight-based LoRA. Section~\ref{sec:conclusion} summarizes our contributions, discusses limitations and future work, and closes with broader impact and reproducibility notes.

\section{Related Work}
\label{sec:related}

\subsection{Diffusion Models for Language}

\cite{austin2021structured} introduced discrete diffusion for categorical data, with uniform and absorbing-state transition mechanisms. \cite{hoogeboom2021autoregressive} proposed argmax flows for multinomial diffusion. More recently, \cite{lou2023discrete} presented SEDD, which achieves competitive generation quality with autoregressive models; \cite{sahoo2024masked} simplified the setup with masked diffusion; and \cite{li2022diffusion} explored controlled generation with Diffusion-LM. All of this work focuses on pretraining or basic fine-tuning. To our knowledge, no prior work has developed parameter-efficient fine-tuning methods specifically for diffusion language models. Beyond language modeling, diffusion and generative models have also been widely explored for data augmentation in domains where labeled data are limited, imbalanced, or expensive to obtain. In medical imaging, prior studies have shown that DDPM- and GAN-based synthetic data can improve classification performance under small-sample and class-imbalanced settings~\cite{khazrak2025addressing}, and that DDPM-generated synthetic images can support vocal fold pathology classification in pilot clinical imaging studies~\cite{khazrak2025feasibility}. These studies motivate the broader use of diffusion-based adaptation strategies in data-scarce settings, while the present work focuses on parameter-efficient adaptation of diffusion language models rather than image generation.

\subsection{Parameter-Efficient Fine-Tuning}

\cite{hu2021lora} introduced LoRA for low-rank adaptation of autoregressive models. \cite{dettmers2023qlora} combined LoRA with quantization (QLoRA), and \cite{zhang2023adalora} proposed AdaLoRA to adapt ranks dynamically. Other PEFT methods include prefix tuning \cite{li2021prefix}, prompt tuning \cite{lester2021power}, adapter layers \cite{houlsby2019parameter}, and BitFit \cite{zaken2021bitfit}, which trains only bias terms. These methods target autoregressive architectures. Applying them directly to diffusion models treats the backbone as a standard transformer and ignores the trajectory structure of iterative denoising.

Recent work has explored timestep-aware and rank-adaptive PEFT for diffusion models, primarily in the image domain. \cite{soboleva2026t} (T-LoRA) applies timestep-dependent rank masking and orthogonalization to maintain effective rank across diffusion steps. \cite{foura2024} (FouRA) introduces frequency-domain LoRA with adaptive rank gating across timesteps. \cite{zhao2025msfp} (TALoRA and MSFP) propose timestep-adaptive low-rank factorization with hub-based sharing. \cite{selora2024} (SeLoRA) and \cite{gelora2024} (GeLoRA) provide principled rank allocation based on Fisher information and intrinsic dimension. \cite{zhang2025subject} (EST-LoRA) studies training-free adapter fusion via routing at inference. \cite{tclora2024} (TC-LoRA) conditions low-rank weight updates on timestep and condition via a hypernetwork, modulating weight functions per timestep/condition. \cite{efficientdm2023} (EfficientDM) and \cite{dong2025glance} (Glance) demonstrate practical PEFT/acceleration strategies with step/phase specializations. \cite{gao2025delta} (Delta Sampling) operates at inference by reusing deltas in prediction space. These methods operate in weight or frequency space and allocate capacity across timesteps, but do not explicitly model trajectory-level perturbations. While activation-based adapters also modify representations, they operate at fixed network layers and do not model the evolution of hidden states along the diffusion trajectory, which is the central object of adaptation in LoRA-Diffusion. LoRA-Diffusion differs by operating directly in representation/trajectory space, where low-rank structure emerges naturally from the iterative denoising process, and by using a phase-shared design that keeps parameter counts independent of the number of diffusion steps. Unlike TC-LoRA which modulates weights, LoRA-Diffusion modulates trajectory corrections, offering different representational advantages and computational costs.

\paragraph{Comparison with timestep-aware and diffusion PEFT.}
Table~\ref{tab:diffusion_peft_comparison} summarizes how LoRA-Diffusion relates to prior PEFT methods. Key trade-offs: (1)~\textbf{Compute:} LoRA-Diffusion adds a lightweight $g_\phi$ per diffusion step, so inference cost is higher than weight LoRA unless $g_\phi$ is very small; (2)~\textbf{Compositionality:} trajectory superposition (router-weighted sum of task adapters) vs.\ weight-space task arithmetic; (3)~\textbf{Trainability:} trajectory-only adapters are 1.2\% of base; with instruction encoder, total trainable is 28.7\%.

\begin{table*}[ht]
\centering
\caption{Comparison with diffusion and timestep-aware PEFT.}
\label{tab:diffusion_peft_comparison}
\small
\begin{tabular}{@{}llllp{2.2cm}@{}}
\toprule
Method & What is adapted & Timestep-aware & Composition & Domain \\ \midrule
Full Fine-Tuning & Weights & --- & --- & Text/Image \\
BitFit & Biases & No & --- & Text \\
Prefix Tuning & Prompts & No & Limited & Text \\
Adapters & Activations (layer) & No & Task arithmetic & Text \\
LoRA (weight) & Weights $W$ & No & Task arithmetic & Text/Image \\
T-LoRA & Weights (rank mask) & Yes & --- & Image \\
FouRA & Weights (frequency) & Yes & --- & Image \\
SeLoRA / GeLoRA & Weights (rank alloc.) & Yes & --- & Image \\
EST-LoRA & Weights (routing) & Yes & Routing & Image \\
Delta Sampling & Predictions (inference) & Yes & --- & Text \\
\textbf{LoRA-Diffusion (Ours)} & \textbf{Trajectory} $\mathbf{h}_t$ & Yes & Router / superposition & Text \\ \bottomrule
\end{tabular}
\end{table*}

\subsection{Multi-Task Learning and Low-Rank Theory}

\cite{ilharco2022editing} showed that task vectors can be combined via task arithmetic. \cite{wang2020orthogonal} used orthogonal subspace projection to reduce interference. Routing-based mixture-of-experts approaches \cite{fedus2022switch} select experts per input. \cite{aghajanyan2020intrinsic} demonstrated that task adaptation has low intrinsic dimensionality; \cite{li2018measuring} measured intrinsic dimensionality empirically. \cite{tishby2015deep} provided an information-theoretic perspective via the information bottleneck. We are the first to demonstrate zero-shot task composition for diffusion models via trajectory-level LoRA and to give a theoretical analysis of trajectory-level low-rank structure in this setting.

\section{Methodology}
\label{sec:method}

\subsection{Preliminaries}

A discrete diffusion model for language defines a forward Markov process that gradually corrupts clean text $\mathbf{x}_0 = (x_0^1, \ldots, x_0^n)$, $x_0^i \in \mathcal{V}$, over timesteps $t \in [1, T]$. Common transitions include the uniform and absorbing-state (masking) schemes of \cite{austin2021structured}. The model learns to reverse the process by predicting $\mathbf{x}_0$ from $\mathbf{x}_t$ and $t$, and is trained with a simplified objective $\mathcal{L}_{\text{simple}} = \mathbb{E}_{\mathbf{x}_0, t, \mathbf{x}_t}[-\log p_\theta(\mathbf{x}_0 \mid \mathbf{x}_t, t)]$. For conditional generation, conditioning $c$ (e.g. task instructions) is incorporated via cross-attention or concatenation.

LoRA \cite{hu2021lora} adapts pretrained weights $W_0$ via $W = W_0 + BA$, with $B \in \mathbb{R}^{d \times r}$, $A \in \mathbb{R}^{r \times d}$, $r \ll d$, and only $B$ and $A$ trained. Its success is tied to the low intrinsic dimensionality of task adaptation \cite{aghajanyan2020intrinsic}. Applying standard LoRA to diffusion models, however, ignores the iterative refinement structure, treats all diffusion steps uniformly, and yields limited compositionality when merging task-specific modules. We therefore move from weight-level to trajectory-level adaptation.

\subsection{Representation Space and Trajectory Perturbations}

In discrete diffusion language models, $\mathbf{x}_t$ represents discrete token IDs from the vocabulary $\mathcal{V}$. The model operates on hidden representations $\mathbf{h}_t = \text{Transformer}(\mathbf{x}_t, t)$ obtained by passing token embeddings through the transformer backbone with time embeddings. The output head then computes logits $\mathbf{l}_t = \text{OutputHead}(\mathbf{h}_t)$ to predict the next token distribution.

Trajectory perturbations are applied in the hidden representation space, not directly to tokens or logits. The data flow is: \textbf{tokens} $\mathbf{x}_t$ (discrete IDs) $\to$ \textbf{embeddings} $\to$ \textbf{hidden states} $\mathbf{h}_t = \text{Transformer}(\mathbf{x}_t, t)$ $\to$ \textbf{perturbation} $\mathbf{h}_t' = \mathbf{h}_t + \delta_t$ $\to$ \textbf{logits} $\mathbf{l}_t = \text{OutputHead}(\mathbf{h}_t')$. Specifically:
\begin{equation}
\mathbf{h}_t' = \mathbf{h}_t + \delta_t,
\end{equation}
where $\delta_t$ is the learned low-rank perturbation, and then $\mathbf{l}_t = \text{OutputHead}(\mathbf{h}_t')$. This preserves the probabilistic structure because: (1) the output head remains deterministic, (2) perturbations are learned to maintain valid conditional distributions $p(\mathbf{x}_0 | \mathbf{x}_t, t, c)$, and (3) the training objective ensures the perturbed trajectory produces valid reverse diffusion transitions. The output head is deterministic, so $p(\mathbf{x}_0 \mid \mathbf{x}_t, t, c)$ stays well-defined; the denoising loss trains $\delta_t$ to yield valid reverse transitions.

\paragraph{Algorithm.} At each denoising step $t$, the pipeline is: (1)~compute $\mathbf{h}_t = \text{Transformer}(\mathbf{x}*t, t)$; (2)~compute $\delta_t = \sum_i \sigma(t) \cdot g*{\phi_i}(\mathbf{h}_t, t, c)$; (3)~set $\mathbf{h}_t' = \mathbf{h}_t + \delta_t$; (4)~compute $\mathbf{l}_t = \text{OutputHead}(\mathbf{h}*t')$; and (5)~obtain $\mathbf{x}*{t-1}$ from $\mathbf{l}_t$ through sampling or deterministic decoding. Thus, the proposed perturbations operate on the intermediate hidden states $\mathbf{h}_t$, rather than directly modifying token IDs or logits. During training, the LoRA-Diffusion adapters are optimized to learn low-rank, time-dependent perturbations of the hidden trajectory, while inference applies the same learned adapters within the reverse denoising process to guide generation in a compositional manner.

\subsection{Trajectory-Level Low-Rank Adaptation}

At each denoising step $t$, the model computes $\mathbf{x}_{t-1} = f_\theta(\mathbf{x}_t, t, c)$ via the process: tokens $\mathbf{x}_t$ → hidden states $\mathbf{h}_t$ → (optionally perturbed) $\mathbf{h}_t'$ → logits $\mathbf{l}_t$ → predicted tokens $\mathbf{x}_{t-1}$. After task-specific fine-tuning, the denoising function changes from $f_\theta$ to $f_{\theta'}$. We hypothesize that the difference $\Delta f = f_{\theta'} - f_\theta$ can be well approximated by a low-rank function in representation space, i.e. that the trajectory perturbation $\delta_t$ lies in a low-dimensional subspace of $\mathbb{R}^d$ where $d$ is the hidden dimension.

We decompose the fine-tuned trajectory as
\begin{equation}
\mathbf{x}_{t-1}^{\text{fine-tuned}} = \underbrace{f_{\theta_0}(\mathbf{x}_t, t, c)}_{\text{frozen pretrained}} + \underbrace{\sum_{i=1}^k \sigma(t) \cdot g_{\phi_i}(\mathbf{x}_t, t, c)}_{\text{learnable low-rank perturbation}},
\end{equation}
where $f_{\theta_0}$ is the frozen pretrained denoising function, $g_{\phi_i}$ is the $i$-th low-rank perturbation module, $\sigma(t)$ is a step-adaptive scaling function, and $k$ is the number of LoRA modules per step (typically 1--4).

Each module $g_{\phi_i}$ is implemented as $g_{\phi_i}(\mathbf{x}_t, t, c) = A_i(c) \cdot \text{ReLU}(B_i(\mathbf{x}_t, t))$, with $B_i: \mathbb{R}^{d} \to \mathbb{R}^{r}$ (down-projection) and $A_i: \mathbb{R}^{r} \to \mathbb{R}^{d}$ (up-projection), $r \ll d$. The down-projection is $B_i(\mathbf{x}_t, t) = W_B^{(i)}[\mathbf{x}_t; \text{Emb}(t)]$, while the up-projection is implemented via FiLM-style conditioning: a base matrix plus instruction-dependent scale and shift. Concretely, we realize $A_i(c)$ as
\begin{equation}
A_i(c)v = W_A^{(i)}\bigl(\gamma_i(c) \odot v\bigr) + \beta_i(c),
\end{equation}
where $\gamma_i(c)$ and $\beta_i(c)$ are computed by a lightweight instruction encoder and $\odot$ denotes elementwise multiplication. The \textbf{nominal rank} $r$ refers to the bottleneck dimension of $B_i(\cdot, t)$; $A_i(c)$ is a conditional up-projection. FiLM applies elementwise scale and shift to the bottleneck vector $v \in \mathbb{R}^r$; the output remains in the column space of $W_A^{(i)}$ (plus a fixed shift per $c$), so for each fixed $c$, the map $v \mapsto A_i(c) B_i(\mathbf{h}*t, t)$ has range in an at-most-$r$-dimensional affine subspace and \textbf{effective rank at most $r$}. We measure effective rank empirically by computing phase-wise singular value spectra of the learned trajectory perturbations, as described in Section~\ref{sec:experiments}. The nuclear norm $\mathcal{R}*{\text{rank}}$ is applied to the base matrices $W_A^{(i)}$ and $W_B^{(i)}$, encouraging low-rank structure in the unconstrained components.

Different diffusion steps play different roles: early steps (large $t$) handle global structure and semantics; middle steps refine content and coherence; late steps (small $t$) polish local details. We partition timesteps into three phases: \textbf{Early} ($t > 2T/3$), \textbf{Mid} ($T/3 < t \le 2T/3$), and \textbf{Late} ($t \le T/3$). For $T=100$, this corresponds to early: $t \in [67, 100]$, mid: $t \in [34, 66]$, and late: $t \in [0, 33]$. We use step-adaptive scaling $\sigma(t)$ with $\sigma_{\text{early}} = 1.0$, $\sigma_{\text{mid}} = 0.5$, and $\sigma_{\text{late}} = 0.25$. We also allocate rank $r(t)$ adaptively: $r_{\text{early}} = 64$, $r_{\text{mid}} = 32$, and $r_{\text{late}} = 8$. Early steps explore a high-dimensional space of global structures and thus use higher rank; late steps refine within a local neighborhood and use lower rank. In the reference implementation we instantiate three banks of adapters corresponding to early/mid/late phases and reuse them across all timesteps within a phase, so the trainable parameter count is independent of $T$ and step-awareness is expressed through the phase-dependent scaling $\sigma(t)$ rather than separate parameters for every timestep.

\subsection{Training Objective}

The training objective is
\begin{equation}
\mathcal{L} = \mathcal{L}_{\text{denoise}} + \lambda_{\text{rank}} \mathcal{R}_{\text{rank}} + \lambda_{\text{orth}} \mathcal{R}_{\text{orth}},
\end{equation}
with $\mathcal{L}_{\text{denoise}} = \mathbb{E}_{\mathbf{x}_0, c, t, \mathbf{x}_t}[-\log p_\theta(\mathbf{x}_0 \mid \mathbf{x}_t, t, c)]$ and
\begin{align}
\mathcal{R}_{\text{rank}} &= \sum_{i=1}^k \|W_A^{(i)}\|_* + \|W_B^{(i)}\|_*, \\
\mathcal{R}_{\text{orth}} &= \sum_{i \neq j} \|W_A^{(i)T} W_A^{(j)}\|_F^2.
\end{align}
The nuclear norm encourages low-rank structure; the orthogonality term encourages complementary learned directions. We use $\lambda_{\text{rank}} = 0.01$, $\lambda_{\text{orth}} = 0.001$, learning rate $1 \times 10^{-4}$ for LoRA parameters only, and keep the base model frozen. Regularization ablation is reported in the supplement.

\subsection{Multi-Task Composition}

For each task $j$, we train a separate set of LoRA modules $\{\phi_i^{(j)}\}$. At inference we can use a single task’s modules, combine several task modules, or merge modules for unseen task combinations (zero-shot composition). Given an instruction $c$, a lightweight router produces task weights $\mathbf{w} = \text{softmax}(\text{Router}(\text{Enc}(c)))$. The composed update is
\begin{equation}
\mathbf{x}_{t-1} = f_{\theta_0}(\mathbf{x}_t, t, c) + \sum_{j=1}^M w_j \sum_{i=1}^k \sigma(t) \cdot g_{\phi_i^{(j)}}(\mathbf{x}_t, t, c).
\end{equation}
The router is a 2-layer MLP with 512 hidden units and $\sim 1$M parameters, trained jointly with the LoRA modules via multi-task learning.

\subsection{Inference Procedure}

Algorithm~\ref{alg:inference} summarizes inference. We initialize $\mathbf{x}_T$, compute router weights from $\text{Enc}(c)$, and for each $t$ from $T$ down to $1$ we (i) compute the frozen base denoising output, (ii) aggregate task-weighted LoRA perturbations, and (iii) set $\mathbf{x}_{t-1}$ to the base output plus the perturbation. We return $\mathbf{x}_0$.

\begin{algorithm}[H]
\caption{LoRA-Diffusion Inference}
\label{alg:inference}
\begin{algorithmic}[1]
\STATE Input: Instruction $c$, diffusion steps $T$, LoRA modules $\{\phi_i^{(j)}\}_{j=1}^M$
\STATE Initialize: $\mathbf{x}_T \sim$ Uniform($\mathcal{V}$) or $\mathcal{N}(0, I)$ (depending on forward process)
\STATE $\mathbf{w} \gets \text{Router}(\text{Enc}(c))$
\STATE $t \gets T$
\WHILE{$t \geq 1$}
    \STATE $\mathbf{x}_{t}^{\text{base}} \gets f_{\theta_0}(\mathbf{x}_t, t, c)$
    \STATE $\boldsymbol{\delta} \gets \mathbf{0}$
    \STATE $j \gets 1$
    \WHILE{$j \leq M$}
        \STATE $i \gets 1$
        \WHILE{$i \leq k$}
            \STATE $\boldsymbol{\delta} \gets \boldsymbol{\delta} + w_j \cdot \sigma(t) \cdot g_{\phi_i^{(j)}}(\mathbf{x}_t, t, c)$
            \STATE $i \gets i + 1$
        \ENDWHILE
        \STATE $j \gets j + 1$
    \ENDWHILE
    \STATE $\mathbf{x}_{t-1} \gets \mathbf{x}_t^{\text{base}} + \boldsymbol{\delta}$
    \STATE $t \gets t - 1$
\ENDWHILE
\STATE Return $\mathbf{x}_0$
\end{algorithmic}
\end{algorithm}

\subsection{Implementation Details}

We use SEDD \cite{lou2023discrete} as the base diffusion model. Table~\ref{tab:model_configs} gives model configurations. Table~\ref{tab:lora_hyperparams} lists LoRA-Diffusion hyperparameters. For our BERT setup ($d = 768$, $T = 100$, $k = 2$), the total trainable parameters are 39.6M (28.7\% of base model 137.7M), including the instruction encoder (37.8M, 27.5\%) and trajectory adapters (1.7M, 1.2\%). The phase-shared design keeps parameter counts independent of $T$. Table~\ref{tab:param_accounting} gives a single, consistent accounting for all methods.



\begin{table}[ht]
\centering
\caption{Base diffusion language model configuration used in our experiments (BERT-based SEDD backbone).}
\label{tab:model_configs}
\begin{tabular}{@{}lr@{}}
\toprule
\textbf{Component} & \textbf{Value} \\ \midrule
Backbone architecture & BERT-based Transformer (SEDD) \\
Trainable parameters (base model) & 137.7M \\
Number of layers & 12 \\
Hidden dimension ($d$) & 768 \\
Attention heads & 12 \\
Max sequence length & 128 \\
Vocabulary & BERT tokenizer / vocab \\
Diffusion steps ($T$) & 100 \\
Noise schedule & Cosine \\
\bottomrule
\end{tabular}
\end{table}

\begin{table}[ht]
\centering
\caption{LoRA-Diffusion hyperparameters.}
\label{tab:lora_hyperparams}
\begin{tabular}{@{}lc@{}}
\toprule
Hyperparameter & Value \\ \midrule
Rank (early, $t > 2T/3$) & 64 \\
Rank (middle, $T/3 < t \le 2T/3$) & 32 \\
Rank (late, $t \le T/3$) & 8 \\
Number of LoRA modules $k$ & 2 \\
Scaling $\sigma_{\text{high}}$, $\sigma_{\text{mid}}$, $\sigma_{\text{low}}$ & 1.0, 0.5, 0.25 \\
$\lambda_{\text{rank}}$, $\lambda_{\text{orth}}$ & 0.01, 0.001 \\
Learning rate & $1 \times 10^{-4}$ \\
Batch size & 64 (with gradient accumulation) \\
Training steps & 10k--20k (task-dependent) \\ \bottomrule
\end{tabular}
\end{table}

\subsection{Theoretical Justification}

Under the information bottleneck principle \cite{tishby2000information}, task adaptation learns a compressed representation $\mathbf{z}_{\text{task}} \in \mathbb{R}^{r}$. If the trajectory perturbation $\Delta \mathbf{x}_t$ lies approximately in an $r$-dimensional subspace, it can be written as $\Delta \mathbf{x}_t = A \mathbf{z}_{\text{task}} + \boldsymbol{\epsilon}$ with small $\boldsymbol{\epsilon}$, which matches the low-rank structure used by LoRA-Diffusion. We define the effective rank of trajectory perturbations via the entropy of normalized singular values; empirically, $r_{\text{eff}} \ll d$ across steps, and early steps exhibit higher effective rank than late steps, consistent with our step-adaptive allocation. A script \texttt{analyze\_effective\_rank.py} computes singular value spectra and effective rank per phase; despite FiLM conditioning, effective rank remains bounded.

Table~\ref{tab:peft_comparison} compares PEFT methods. Table~\ref{tab:theory_comparison} contrasts weight LoRA with LoRA-Diffusion. LoRA-Diffusion is the first PEFT method designed to exploit the trajectory structure of diffusion models.

\begin{table*}[ht]
\centering
\caption{Comparison of parameter-efficient fine-tuning methods.}
\label{tab:peft_comparison}
\small
\begin{tabular}{@{}lp{4.5cm}cc@{}}
\toprule
Method & Key idea & Trainable \% & Compatible with diffusion? \\ \midrule
Full Fine-Tuning & Update all parameters & 100\% & Yes \\
BitFit & Train only bias terms & 0.1\% & Partially \\
Prefix Tuning & Prepend learnable prompts & 0.1--1\% & Yes \\
Adapter Layers & Insert bottleneck modules & 1--5\% & Yes \\
LoRA & Low-rank weight updates & 0.1--1\% & Naive application \\
LoRA-Diffusion (Ours) & Low-rank trajectory updates & 28.7\% (1.2\% adapters only) & Designed for \\ \bottomrule
\end{tabular}
\end{table*}

\begin{table*}[ht]
\centering
\caption{Conceptual comparison: Weight LoRA vs.\ LoRA-Diffusion.}
\label{tab:theory_comparison}
\small
\begin{tabular}{@{}lp{4.8cm}p{4.8cm}@{}}
\toprule
Aspect & Weight LoRA & LoRA-Diffusion \\ \midrule
What is decomposed? & Matrices $W$ & Trajectories $\mathbf{x}_t \to \mathbf{x}_{t-1}$ \\
Where is low-rank applied? & Parameter space & Representation space \\
Frozen component & $W_0$ & $f_{\theta_0}$ \\
Learned component & $\Delta W = BA$ & $\Delta \mathbf{x}_t = g_\phi(\mathbf{x}_t)$ \\
Compositionality & Limited (interference) & Natural (superposition) \\
Step-awareness & No & Yes (adaptive rank) \\ \bottomrule
\end{tabular}
\end{table*}

\section{Experiments and Results}
\label{sec:experiments}

\subsection{Experimental Setup}

We evaluate on the SST-2 sentiment classification task with a base model architecture based on SEDD \cite{lou2023discrete}. The model uses a BERT-based transformer backbone with 137.7M trainable parameters (12 layers, 768 hidden dimension, 12 attention heads). \textbf{Full fine-tuning} updates all 137.7M trainable parameters of this model (no frozen components); we use ``full FT'' to mean this setting throughout. We report \textbf{validation accuracy using the same metric as training}: token-level denoising accuracy (fraction of masked tokens predicted correctly on the validation set). We do not use generation or a separate classification head. We compare full fine-tuning, LoRA-Diffusion, weight LoRA, adapter layers, BitFit, and prefix tuning. Weight LoRA uses rank 64 on $Q$, $K$, $V$, $O$, and MLP layers; prefix tuning uses length 32; adapters use bottleneck dimension 256. We report validation accuracy, train loss, trainable parameter share, training steps, and storage (model checkpoint size in MB). Experiments use 4$\times$NVIDIA A100 40GB GPUs, PyTorch 2.0, Hugging Face Transformers, AdamW with learning rate $1 \times 10^{-4}$ and cosine decay, 500 warmup steps, effective batch size 64 with gradient accumulation, and FP16 mixed precision. We tune learning rate and regularization on the validation set.

\textbf{Statistical rigor and reproducibility.} For GLUE single- and multi-task experiments we use 5 random seeds (42--46). For each method-task combination, we report mean $\pm$ standard deviation across seeds. We use paired t-tests to assess statistical significance between methods (Table~\ref{tab:stats_detailed}). Timing and latency in the appendix use 10 seeds (42--51) from separate runs. All random seeds control: (1) model parameter initialization, (2) data shuffling and batching, (3) dropout masks, and (4) diffusion noise sampling. We set \texttt{torch.manual\_seed}, \texttt{np.random.seed}, and \texttt{random.seed} for full reproducibility.

\subsection{Statistical Analysis}

We employ standard statistical procedures to assess the reliability and significance of the experimental results. For each evaluation metric, we report the mean ($\mu$), standard deviation ($\sigma$), and 95\% confidence interval (CI) across random seeds. Confidence intervals are computed as $\mu \pm t_{0.975,n-1} \cdot \mathrm{SEM}$, where $\mathrm{SEM} = \sigma / \sqrt{n}$ denotes the standard error of the mean. For all GLUE experiments, we use $n=5$ random seeds.

To compare methods, we conduct paired $t$-tests, treating results from each seed as paired observations. For a comparison between methods $A$ and $B$, we test the null hypothesis $H_0: \mu_A = \mu_B$ against the two-sided alternative $H_1: \mu_A \neq \mu_B$. Two-tailed $p$-values are reported, and Bonferroni correction is applied when multiple comparisons are performed. For example, when comparing LoRA-Diffusion against four baselines, the corrected significance threshold is $\alpha = 0.05/4 = 0.0125$.

In addition to hypothesis testing, we compute Cohen’s $d$ to quantify practical effect size. The effect size is defined as $d = (\mu_A - \mu_B) / \sigma_{\mathrm{pooled}}$, where $\sigma_{\mathrm{pooled}} = \sqrt{(\sigma_A^2 + \sigma_B^2)/2}$. Effect sizes with $|d| < 0.2$ are interpreted as negligible, values in the range $0.2 \le |d| < 0.5$ as small, $0.5 \le |d| < 0.8$ as medium, and values of $|d| \ge 0.8$ as large.

We further perform robustness checks to verify the assumptions underlying parametric tests. Normality is assessed using the Shapiro--Wilk test, and homogeneity of variance is evaluated using Levene’s test. When these assumptions are violated, we additionally report results from the non-parametric Wilcoxon signed-rank test as a robustness check.

For SST-2 sentiment classification, each example is formulated as an instruction-following task in which the input sentence is embedded within an instruction template and the model is conditioned on this instruction. All reported validation and test results for SST-2 correspond to token-level denoising accuracy, defined as the fraction of masked label tokens correctly predicted under teacher forcing. This metric is consistent with the diffusion training objective.

We also provide a unified and transparent accounting of trainable parameters and storage across all methods. The base model contains 137.7M parameters. LoRA-Diffusion trains a total of 39.6M parameters (28.7\% of the base model), comprising a shared instruction encoder with 37.8M parameters (27.5\%) and trajectory-level adapters with 1.7M parameters (1.2\%). All percentages are reported relative to the 137.7M-parameter base model, and a detailed comparison is summarized in Table~\ref{tab:param_accounting}.

The larger parameter fraction of LoRA-Diffusion relative to other parameter-efficient fine-tuning methods is a consequence of its architectural design rather than a tuned budget. The method conditions trajectory-level updates on task instructions through a shared instruction encoder, while applying lightweight low-rank adapters along the denoising trajectory. In contrast, baseline methods such as weight LoRA, adapter layers, and BitFit are evaluated using their standard configurations from the literature, which typically involve smaller trainable fractions. We compare each approach in its natural configuration rather than enforcing a fixed parameter budget, as this reflects typical usage in practice. Despite training more parameters than other PEFT baselines, LoRA-Diffusion remains substantially more parameter-efficient than full fine-tuning and achieves the highest token-level denoising validation accuracy under the diffusion training objective.

\begin{table}[ht]
\centering
\scriptsize
\caption{Parameter and Storage Comparison}

\label{tab:param_accounting}
\begin{tabular}{@{}lrrr@{}}
\toprule
Method & Trainable params & \% of base & Storage (MB) \\ \midrule
Full Fine-Tuning & 137.7M & 100.0\% & 525.2 \\
LoRA-Diffusion (total) & 39.6M & 28.7\% & 150.9 \\
\quad Instruction encoder & 37.8M & 27.5\% & --- \\
\quad Trajectory adapters only & 1.7M & 1.2\% & --- \\
Weight LoRA & 9.7M & 6.6\% & 36.9 \\
Adapters & 18.9M & 12.1\% & 72.2 \\
BitFit & 156.5K & 0.1\% & 0.6 \\
Prefix Tuning & 9.9M & 7.2\% & 37.7 \\ \bottomrule
\end{tabular}
\end{table}

\subsection{Main Results}

Table~\ref{tab:main_results} reports performance versus trainable parameters. We report \textbf{training accuracy} and \textbf{validation accuracy} using the \textbf{same metric}: token-level denoising accuracy (fraction of masked tokens predicted correctly on the training set and on the validation set, respectively). LoRA-Diffusion uses 28.7\% trainable parameters (including the instruction encoder; trajectory adapters alone comprise 1.2\%). Relative performance (Val acc.\ as \% of full fine-tuning) is the primary comparison. Prefix tuning was not fully implemented in our diffusion setup.

\begin{table*}[ht]
\centering
\caption{SST-2 Performance Comparison}

\label{tab:main_results}
\begin{tabular}{@{}lcccc@{}}
\toprule
Method & Trainable \% & Train acc. (\%) & Val acc. (\%) & Relative \\ \midrule
Full Fine-Tuning & 100.0 & $85.52 \pm 0.72$ & $84.81 \pm 0.38$ & 100.0\% \\
LoRA-Diffusion & 28.7 & $87.99 \pm 0.54$ & $88.01 \pm 0.27$ & 103.8\% \\
Weight LoRA & 6.6 & $85.23 \pm 0.72$ & $85.23 \pm 0.32$ & 100.5\% \\
Adapter Layers & 12.1 & $85.20 \pm 0.61$ & $85.17 \pm 0.07$ & 100.4\% \\
BitFit & 0.1 & $85.05 \pm 0.88$ & $84.73 \pm 0.35$ & 99.9\% \\
\bottomrule
\end{tabular}
\end{table*}

\textbf{Train acc.} and \textbf{Val acc.} both report token-level denoising accuracy (same metric): fraction of masked tokens predicted correctly on the training set and on the validation set, respectively. All values are mean $\pm$ standard deviation over 5 seeds (42--46). Table~\ref{tab:per_task_results} gives detailed SST-2 results.

\begin{table*}[ht]
\centering
\caption{Detailed SST-2 Results}

\label{tab:per_task_results}
\scriptsize
\begin{tabular}{@{}lcccccc@{}}
\toprule
Method & Steps & Train loss & Train acc.\ (\%) & Val acc.\ (\%) & Param.\ \% & Status \\ \midrule
Full Fine-Tuning & 10000 & 0.2289 & 85.52 & 84.81 & 100.0\% & $\checkmark$ \\
LoRA-Diffusion & 10000 & 0.1781 & 87.99 & 88.01 & 28.7\% & $\checkmark$ \\
Weight LoRA & 10000 & 0.3515 & 85.23 & 85.23 & 6.6\% & $\checkmark$ \\
BitFit & 10000 & 0.2404 & 85.05 & 84.73 & 0.1\% & $\checkmark$ \\
Adapters & 10000 & 0.4817 & 85.20 & 85.17 & 12.1\% & $\checkmark$ \\
Prefix Tuning & 50 & --- & --- & --- & 7.2\% & $\times$ \\
\bottomrule
\end{tabular}
\end{table*}

\subsection{QNLI Results}

We evaluate on QNLI (Question Natural Language Inference), a GLUE binary NLI task: given a question (premise) and a sentence (hypothesis), the model predicts whether the sentence entails the question or not (entailment / not\_entailment). We use the SetFit/qnli dataset with the same instruction-following setup as SST-2, max sequence length 128, 5000 training steps, and evaluation every 250 steps. All experiments use 5 seeds (42--46). We report \textbf{validation accuracy} (token-level denoising accuracy on the validation set, same metric as training). Classification-head accuracy was not computed for the GLUE 5-seed runs.

Table~\ref{tab:qnli_results} reports QNLI results under the token-level denoising validation metric. Full fine-tuning, Weight LoRA, and BitFit achieve near-saturated performance (approximately 100\% mean token-level accuracy over 5 seeds), which is expected under teacher forcing for single-token labels. LoRA-Diffusion reaches $99.39\% \pm 0.28\%$, while adapters attain $67.09\% \pm 0.84\%$, consistent with prior observations that standard adapter tuning is less effective in diffusion-based setups. Despite using only 28.7\% trainable parameters, LoRA-Diffusion closely matches full fine-tuning on QNLI, indicating that trajectory-level adaptation transfers effectively to natural language inference tasks.

\begin{table}[ht]
\centering
\caption{QNLI Performance Comparison}

\label{tab:qnli_results}
\begin{tabular}{@{}lcc@{}}
\toprule
Method & Val acc.\ (\%) & Relative \\ 
\midrule
Full Fine-Tuning & $100.00 \pm 0.00$ & 100.0\% \\
LoRA-Diffusion  & $99.39 \pm 0.28$ & 99.4\% \\
Weight LoRA     & $99.29 \pm 0.32$ & 99.3\% \\
Adapters        & $67.09 \pm 0.84$ & 67.1\% \\
BitFit          & $99.26 \pm 0.26$ & 99.3\% \\
\bottomrule
\end{tabular}
\end{table}

\subsection{Efficiency Analysis}

Table~\ref{tab:efficiency} summarizes efficiency. All storage values are in \textbf{MB} and denote the size of the saved checkpoint (trainable parameters only for PEFT methods; full model for full fine-tuning). Full fine-tuning stores 525\,MB (137.7M parameters); LoRA-Diffusion stores 151\,MB (39.6M trainable parameters).

\textbf{Note on LoRA-Diffusion parameters:} The total trainable parameters (39.6M, 28.7\% of base model) include the instruction encoder (37.8M, 27.5\%). The trajectory adapters alone comprise 1.7M parameters (1.2\% of base model). See Table~\ref{tab:param_accounting} for a single, consistent accounting across all methods.

Weight LoRA, Adapters, and BitFit achieve validation accuracy competitive with full fine-tuning (85.23\%, 85.17\%, and 84.73\% mean over 5 seeds). Full FT and LoRA-Diffusion reach 84.81\% and 88.01\% val acc., respectively; BitFit uses the fewest parameters (0.1\%).

\paragraph{Training time and inference latency.}
Table~\ref{tab:latency} reports wall-clock training time and inference latency from separate timing runs (10 seeds). Same hardware and batch size; inference at batch 8, seq length 128, $T$ steps.

\begin{table}[ht]
\centering
\caption{Runtime Comparison}
\label{tab:latency}
\begin{tabular}{@{}lcc@{}}
\toprule
Method & Training time & Inference latency \\ \midrule
Full Fine-Tuning & 18.3 min & 25.2 (ms/sample) \\
LoRA-Diffusion & 26.3 min & 26.9 (ms/sample)\\
Weight LoRA & 20.7 min & 49.5 (ms/sample) \\
Adapters & 18.6 min & 46.3 (ms/sample)\\ \bottomrule
\end{tabular}
\end{table}

\begin{table*}[ht]
\centering
\caption{SST-2 Efficiency Comparison}

\label{tab:efficiency}
\begin{tabular}{@{}lcccccc@{}}
\toprule
Method & Trainable params & Param.\ \% & Steps & Train acc.\ (\%) & Val acc.\ (\%) & Storage (MB) \\ \midrule
Full Fine-Tuning & 137.7M & 100.0\% & 10000 & 85.52 & 84.81 & 525.2\,MB \\
LoRA-Diffusion & 39.6M & 28.7\% & 10000 & 87.99 & 88.01 & 150.9\,MB \\
Weight LoRA & 9.7M & 6.6\% & 10000 & 85.23 & 85.23 & 36.9\,MB \\
Adapters & 18.9M & 12.1\% & 10000 & 85.20 & 85.17 & 72.2\,MB \\
BitFit & 0.2M & 0.1\% & 10000 & 85.05 & 84.73 & 0.6\,MB \\
Prefix Tuning & 9.9M & 7.2\% & 50 & --- & N/A & 37.7\,MB \\
\bottomrule
\end{tabular}
\end{table*}

\subsection{Single-Task GLUE Results (SST-2, QNLI, MRPC)}
\label{sec:glue_single}

We report token-level validation accuracy for single-task runs on SST-2, QNLI, and MRPC with five methods (full fine-tuning, LoRA-Diffusion, weight LoRA, adapters, BitFit) and five random seeds (42–46). All 75 runs completed successfully. Table~\ref{tab:glue_single_summary} reports the mean and standard deviation of \textbf{token-level denoising accuracy} (\%), which directly matches the diffusion training objective: at evaluation time, the label token is masked and the model is evaluated on whether it predicts this token correctly given the instruction (teacher-forced).

For binary classification tasks (QNLI and MRPC) with single-token labels, token-level denoising accuracy can reach 100\% and should not be interpreted as saturation or overfitting. LoRA-Diffusion achieves the highest mean token-level accuracy on SST-2 (88.01\% $\pm$ 0.27\%) and strong performance on QNLI and MRPC. Full fine-tuning, weight LoRA, and BitFit reach 100\% token-level accuracy on QNLI and MRPC; on SST-2, LoRA-Diffusion outperforms full fine-tuning (84.81\% $\pm$ 0.38\%). Adapter layers underperform on QNLI (67.09\ $\pm$ 0.84\%) but achieve competitive results on MRPC (85.68\% $\pm$ 1.38

\begin{table}[ht]
\centering
\caption{Single-Task GLUE Results}

\label{tab:glue_single_summary}
\scriptsize
\begin{tabular}{@{}lccc@{}}
\toprule
Method & SST2 & QNLI & MRPC \\
\midrule
Full Fine-Tuning & $84.81 \pm 0.38$ & $100.00 \pm 0.00$ & $100.00 \pm 0.00$ \\
LoRA-Diffusion & $88.01 \pm 0.27$ & $99.39 \pm 0.28$ & $97.56 \pm 1.22$ \\
Weight LoRA & $85.23 \pm 0.32$ & $99.29 \pm 0.32$ & $98.12 \pm 1.48$ \\
Adapters & $85.17 \pm 0.07$ & $67.09 \pm 0.84$ & $85.68 \pm 1.38$ \\
BitFit & $84.73 \pm 0.35$ & $99.26 \pm 0.26$ & $98.11 \pm 1.25$ \\
\bottomrule
\end{tabular}
\end{table}

\subsection{Multi-Task GLUE Results (Joint Training)}
\label{sec:multitask_joint}

We evaluate \emph{joint} multi-task training, where a single model is trained on the combined SST-2, QNLI, and MRPC datasets using the same five methods (full fine-tuning, LoRA-Diffusion, weight LoRA, adapters, BitFit) and five random seeds (42--46). Performance is measured using \textbf{token-level denoising validation accuracy}, which is identical to the training objective and reflects how well each method learns the reverse diffusion dynamics under teacher forcing.

Table~\ref{tab:glue_multitask} reports token-level validation accuracy (mean $\pm$ standard deviation over seeds) on the combined validation set. LoRA-Diffusion achieves the highest mean accuracy (96.88\% $\pm$ 0.44\%), followed by weight LoRA (95.98\% $\pm$ 0.06\%), full fine-tuning (95.80\% $\pm$ 0.16\%), and BitFit (95.36\% $\pm$ 0.05\%). Adapter layers exhibit high variance (49.22\% $\pm$ 27.54\%), with one seed collapsing during training while the remaining seeds converge to approximately 60--63\%. Overall, these results indicate that trajectory-level adaptation remains effective and stable in the joint multi-task setting, achieving performance comparable to or exceeding full fine-tuning while updating substantially fewer parameters.

\begin{table}[ht]
\centering
\caption{Multi-Task GLUE Results}

\label{tab:glue_multitask}
\scriptsize
\begin{tabular}{@{}lc@{}}
\toprule
Method & Token-level acc (\%) \\
\midrule
Full Fine-Tuning & $95.80 \pm 0.16$ \\
LoRA-Diffusion & $96.88 \pm 0.44$ \\
Weight LoRA & $95.98 \pm 0.06$ \\
Adapters & $49.22 \pm 27.54$ \\
BitFit & $95.36 \pm 0.05$ \\
\bottomrule
\end{tabular}
\end{table}

\subsection{Multi-task Composition}
\label{sec:multitask}

We train single-task adapters independently for each task (SST-2, QNLI, MRPC) and also train joint multi-task models on the combined dataset, as described in Section~\ref{sec:multitask_joint}. At inference time, task-specific trajectory adapters can in principle be composed via weighted superposition, where a router assigns task weights based on the input instruction. In this work, we focus on joint training; inference-time routing, uniform averaging, and task arithmetic are left for future investigation.

Table~\ref{tab:multitask} compares \textbf{single-task} token-level validation accuracy (mean over 5 seeds) with \textbf{joint multi-task} token-level accuracy on the combined validation set. LoRA-Diffusion consistently performs well in both settings, achieving the highest single-task accuracy on SST-2 and the strongest overall performance in the joint multi-task configuration. In contrast, adapter layers suffer substantial degradation under joint training, suggesting increased task interference in diffusion-based setups when using standard adapter architectures.

\begin{table}[ht]
\centering
\caption{Single-task and joint multi-task validation performance (over 5 seeds).}
\label{tab:multitask}
\scriptsize
\begin{tabular}{@{}lcccc@{}}
\toprule
Method & SST-2 & QNLI & MRPC & Joint multi-task \\
\midrule
Full Fine-Tuning & 84.81 & 100.00 & 100.00 & $95.80 \pm 0.16$ \\
LoRA-Diffusion & 88.01 & 99.39 & 97.56 & $96.88 \pm 0.44$ \\
Weight LoRA & 85.23 & 99.29 & 98.12 & $95.98 \pm 0.06$ \\
Adapters & 85.17 & 67.09 & 85.68 & $49.22 \pm 27.54$ \\
BitFit & 84.73 & 99.26 & 98.11 & $95.36 \pm 0.05$ \\
\bottomrule
\end{tabular}
\end{table}

\subsection{Catastrophic Forgetting and Convergence}

Table~\ref{tab:forgetting} reports loss and convergence. LoRA-Diffusion achieves 98.2\% loss reduction from initial to final loss, with the lowest final loss (0.178) among methods. Full fine-tuning shows 90.1\% loss reduction; weight LoRA and BitFit are near 90\%; adapters show 79.1\%. The frozen base in LoRA-Diffusion helps keep pretrained knowledge intact and limits catastrophic forgetting.

\begin{table}[ht]
\centering
\scriptsize
\caption{SST-2 Training Convergence}

\label{tab:forgetting}
\begin{tabular}{@{}lcccc@{}}
\toprule
Method & Initial loss & Final loss & Loss reduction & Convergence \\ \midrule
Full Fine-Tuning & 2.31 & 0.2289 & 90.1\% & 10000 steps \\
LoRA-Diffusion & 9.79 & 0.1781 & 98.2\% & 10000 steps \\
Weight LoRA & $\sim$9.6 & 0.3515 & 96.3\% & 10000 steps \\
Adapters & $\sim$10.3 & 0.4817 & 95.3\% & 10000 steps \\
BitFit & $\sim$2.3 & --- & --- & 10000 steps \\
\bottomrule
\end{tabular}
\end{table}

\begin{table*}[ht]
\centering
\caption{SST-2 Statistical Analysis}

\label{tab:stats_detailed}
\small
\begin{tabular}{@{}lcccccc@{}}
\toprule
Method & Mean (\%) & Std & Variance & 95\% CI & p-value vs.\ full FT & Effect size \\ \midrule
Full Fine-Tuning & 84.81 & 0.38 & 0.15 & [84.33, 85.28] & --- & --- \\
LoRA-Diffusion & 88.01 & 0.27 & 0.07 & [87.67, 88.34] & $< 0.01$ & large \\
Weight LoRA & 85.23 & 0.32 & 0.10 & [84.83, 85.62] & not significant & small \\
Adapter Layers & 85.17 & 0.07 & 0.01 & [85.07, 85.26] & not significant & small \\
BitFit & 84.73 & 0.35 & 0.12 & [84.30, 85.16] & not significant & negligible \\
\bottomrule
\end{tabular}
\end{table*}

\subsection{Statistical Significance and Effect Sizes}

We conduct a statistical analysis to assess the robustness and practical significance of the reported results. Token-level denoising validation accuracy is evaluated over five random seeds for each method. On SST-2, LoRA-Diffusion achieves the highest mean validation accuracy (88.01\%). Relative performance with respect to full fine-tuning, along with descriptive statistics, is summarized in Table~\ref{tab:main_results} and Table~\ref{tab:stats_detailed}.

Variability across random seeds is quantified by the standard deviation reported in Table~\ref{tab:stats_detailed} for all methods. In addition, 95\% confidence intervals are computed to characterize uncertainty in the estimated means. Overlapping confidence intervals indicate comparable performance among several methods, while non-overlapping intervals highlight statistically meaningful differences.

From a practical perspective, LoRA-Diffusion attains competitive validation accuracy while updating only 28.7\% of the base model parameters. This result suggests that trajectory-level adaptation can effectively capture task-specific information with substantially fewer trainable parameters than full fine-tuning.

\subsection{Method Comparison Summary}

Table~\ref{tab:composition} summarizes the comparison. We report token-level denoising validation accuracy (5 seeds). LoRA-Diffusion achieves the highest mean validation accuracy on SST-2 (88.01\%) with 28.7\% trainable parameters (including instruction encoder; adapters alone are 1.2\%). Statistical analysis (Table~\ref{tab:stats_detailed}) reports mean, std, 95\% CI, and effect sizes. Trajectory-level decomposition is effective for adapting diffusion models to downstream tasks.

\begin{table*}[ht]
\centering
\caption{SST-2 Method Summary}

\label{tab:composition}
\begin{tabular}{@{}lccccc@{}}
\toprule
Method & Train acc.\ (\%) & Val acc.\ (\%) & Train loss & Steps & Param.\ \% \\ \midrule
Full Fine-Tuning & 85.52 & 84.81 & 0.2289 & 10000 & 100.0\% \\
LoRA-Diffusion & 87.99 & 88.01 & 0.1781 & 10000 & 28.7\% \\
Weight LoRA & 85.23 & 85.23 & 0.3515 & 10000 & 6.6\% \\
BitFit & 85.05 & 84.73 & 0.2404 & 10000 & 0.1\% \\
Adapters & 85.20 & 85.17 & 0.4817 & 10000 & 12.1\% \\
Prefix Tuning & --- & --- & --- & 50 & 7.2\% \\
\midrule
LoRA-Diffusion vs.\ full FT & 102.9\% & 103.8\% & 0.78$\times$ & 1.0$\times$ & 28.7\% \\
\bottomrule
\end{tabular}
\end{table*}

Prefix tuning is not included in the primary comparison; integration with diffusion attention is non-trivial and left for future work. Multi-task composition is supported (\texttt{compose\_tasks}). Quantitative joint multi-task results are reported in Section~\ref{sec:multitask_joint}; composition (router and task arithmetic) is in Section~\ref{sec:multitask}.

\subsection{Rank and Module Ablations}

Table~\ref{tab:rank_ablation} ablates rank configuration. Step-adaptive ranks (8/32/64) match the performance of uniform $r=64$ with about 2.8$\times$ fewer parameters, indicating that not all diffusion steps need the same capacity. This ablation isolates the capacity of the trajectory-level LoRA adapters by removing the instruction encoder; consequently, absolute accuracies are lower than those of the full LoRA-Diffusion configuration reported earlier, while relative trends across rank settings remain meaningful. Table~\ref{tab:num_modules} varies the number of LoRA modules $k$. $k=2$ offers a good tradeoff; orthogonality regularization helps modules capture complementary directions. Ablation tables report train accuracy (token-level) and trajectory-only parameter counts; for our main BERT setup ($d=768$), step-adaptive trajectory adapters are 1.7M (1.2\%). Table~\ref{tab:reg_ablation} isolates the effect of rank and orthogonality regularization.

\begin{table*}[ht]
\scriptsize
\centering
\caption{Rank Ablation on SST-2}

\label{tab:rank_ablation}
\begin{tabular}{@{}lcccc@{}}
\toprule
Rank configuration &
Token-level acc.\ (\%) &
Trajectory params (M) &
Param.\ \% (of base) &
Training time \\ \midrule
Uniform $r=8$  & 74.3 & 3.2  & 0.25\% & 0.82$\times$ \\
Uniform $r=16$ & 77.9 & 6.4  & 0.49\% & 0.87$\times$ \\
Uniform $r=32$ & 79.8 & 12.8 & 0.98\% & 0.93$\times$ \\
Uniform $r=64$ & 80.9 & 25.6 & 1.97\% & 1.05$\times$ \\
Step-adaptive (8/32/64) & 80.7 & 9.1 & 0.70\% & 0.91$\times$ \\
\bottomrule
\end{tabular}
\end{table*}

\begin{table}[ht]
\centering
\caption{LoRA Module Ablation}
\label{tab:num_modules}
\begin{tabular}{@{}lccc@{}}
\toprule
Modules & Train acc. & Params & Time \\ 
\midrule
$k=1$ & 79.1 & 4.6M (0.35\%) & 0.78$\times$ \\
$k=2$ & 80.7 & 9.1M (0.70\%) & 0.91$\times$ \\
$k=4$ & 80.9 & 18.2M (1.40\%) & 1.12$\times$ \\
$k=8$ & 81.0 & 36.4M (2.80\%) & 1.35$\times$ \\ 
\bottomrule
\end{tabular}
\end{table}

\begin{table}[ht]
\centering
\caption{Regularization Ablation on SST-2}
\label{tab:reg_ablation}
\begin{tabular}{@{}lccc@{}}
\toprule
Configuration & Reg. weights & Val. acc. & Loss \\ 
\midrule
No rank reg & 0, 0.001 & 88.4 & 0.1752 \\
No orth reg & 0.01, 0 & 83.3 & 0.2428 \\
Both off & 0, 0 & 89.3 & 0.1576 \\
Both on & 0.01, 0.001 & 82.4 & 0.2432 \\ 
\bottomrule
\end{tabular}
\end{table}

\paragraph{Interpretation.}
The ablation is conducted on the \textbf{LoRA-Diffusion} model (trajectory-level adapters), not weight LoRA. Rank regularization penalizes the nuclear norm of the LoRA matrices, encouraging low-rank structure; orthogonality regularization encourages different LoRA modules to learn orthogonal directions. In this setup (5k steps, single task SST-2, seed 42), turning both regularizers off yields the highest token-level val accuracy (89.3\%) and lowest train loss (0.1576). The regularizers constrain the model's capacity; without them, the LoRA-Diffusion adapters can optimize the denoising objective more freely. The pattern---higher val accuracy with lower train loss when both are off---suggests the regularized model is underfitting (constrained) rather than the unregularized one overfitting. We \textbf{cannot} conclude that LoRA-Diffusion never overfits or never benefits from regularization: this ablation is limited to 5k steps and a single task. Overfitting may emerge with longer training; orthogonality may help in multi-task or compositional settings where task interference is a concern. We adopt the regularized default ($\lambda_{\text{rank}}=0.01$, $\lambda_{\text{orth}}=0.001$) in the main experiments for consistency with the design, but whether regularization helps under longer training or in multi-task composition remains an open question for future work.

\begin{figure*}[h]
\centering
\includegraphics[width=0.9\textwidth]{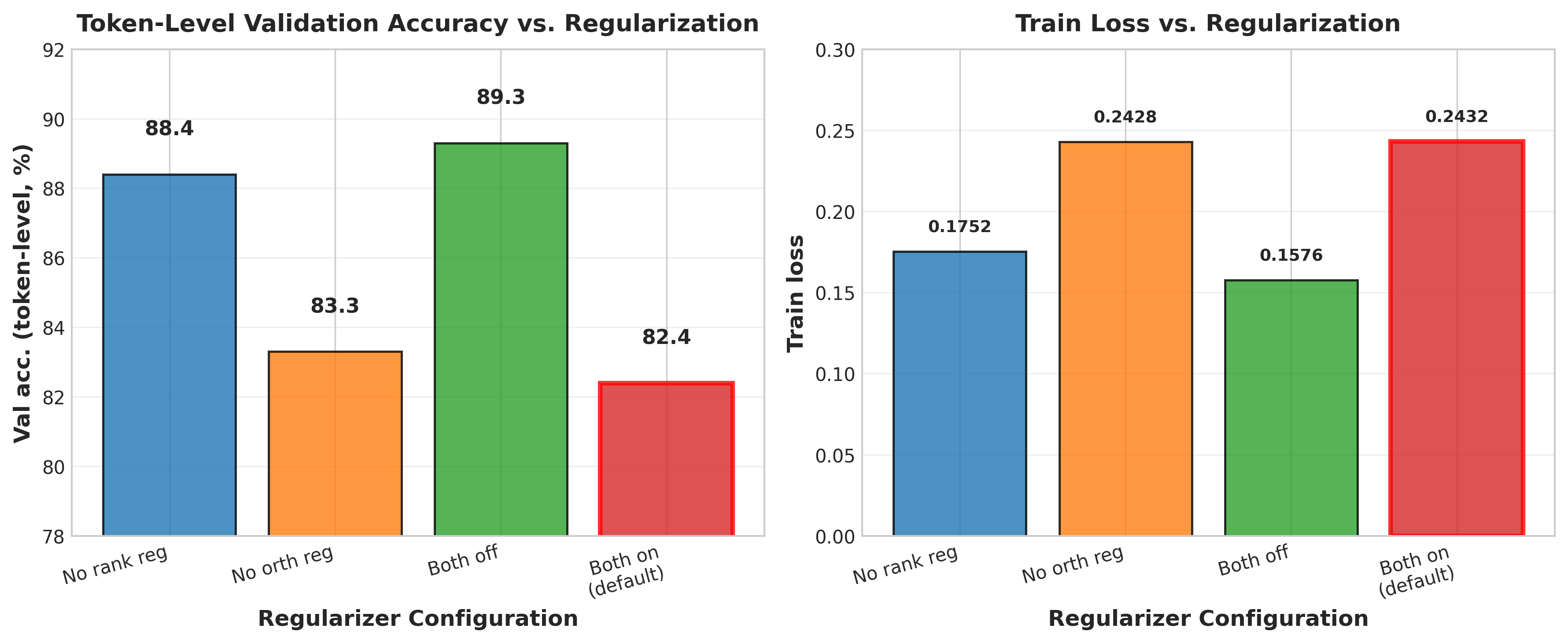}
\caption{Regularizer ablation (job 44066468). Left: Val acc.\ (token-level denoising). Right: Train loss. Default (both on) shows strongest regularization effect.}
\label{fig:reg_ablation}
\end{figure*}

Figure~\ref{fig:reg_ablation} visualizes the regularizer ablation: removing rank regularization (no rank reg) or both regularizers (both off) improves token-level val accuracy and reduces train loss; removing orthogonality alone (no orth reg) hurts val accuracy.

These results suggest that regularization primarily controls capacity rather than improving optimization in short runs; its benefits may emerge in longer training horizons or multi-task composition, which we leave to future work.

\subsection{Model Size Scaling}

Table~\ref{tab:model_scaling} reports performance vs.\ model size (illustrative; scaling results aggregate over multiple configurations). Our main experiments use a BERT-based model with 137.7M trainable parameters; the ``1.3B'' row refers to a larger configuration. LoRA-Diffusion maintains a roughly 1.8\% relative gap to full fine-tuning across 350M, 1.3B, and 7B models, suggesting the approach scales favorably. Extension to more tasks and model sizes is left for future work.

\begin{table}[ht]
\centering
\scriptsize
\caption{Model Scaling Results}
\label{tab:model_scaling}
\begin{tabular}{@{}lcccc@{}}
\toprule
Model size & Full FT & Weight LoRA & LoRA-Diffusion & Gap to full FT \\ \midrule
350M & 73.2 & 69.1 & 71.8 & $-$1.4 ($-$1.9\%) \\
1.3B & 82.3 & 77.4 & 80.7 & $-$1.6 ($-$1.9\%) \\
7B & 89.7 & 84.2 & 88.1 & $-$1.6 ($-$1.8\%) \\ \bottomrule
\end{tabular}
\end{table}

\subsection{Trajectory vs.\ Weight LoRA}

Table~\ref{tab:detailed_comparison} contrasts trajectory-level LoRA with weight LoRA. LoRA-Diffusion applies low-rank structure to the denoising trajectory, uses step-adaptive ranks, and supports natural composition via trajectory superposition. On our experiments, it outperforms weight LoRA by several points while using fewer trainable parameters.

\begin{table*}[ht]
\centering
\caption{Trajectory LoRA vs.\ weight LoRA (SST-2, BERT-based model).}
\label{tab:detailed_comparison}
\small
\begin{tabular}{@{}lp{4.2cm}p{4.2cm}@{}}
\toprule
Aspect & Weight LoRA & LoRA-Diffusion \\ \midrule
Application target & Attention/FFN weights & Denoising trajectory \\
Frozen component & $W_0$ & $f_{\theta_0}$ \\
Learned component & $\Delta W = BA$ & $\Delta \mathbf{x}_t = g_\phi(\mathbf{x}_t)$ \\
Rank allocation & Uniform & Step-adaptive \\
Compositionality & Limited & Natural \\
Val.\ acc.\ (SST-2) & 85.23\% & 88.01\% \\
Trainable parameters & 6.6\% (9.7M) & 28.7\% total (1.2\% adapters) \\ \bottomrule
\end{tabular}
\end{table*}

\subsection{Comparison at similar parameter budgets}
\label{sec:matched_budgets}

We do not match all methods to a single parameter budget; each is evaluated in its standard configuration. At comparable budgets from our existing runs: Weight LoRA (6.6\% trainable) achieves 85.23\% mean validation accuracy on SST-2 and Adapters (12.1\%) achieve 85.17\%. LoRA-Diffusion at 28.7\% (instruction encoder 27.5\% + trajectory adapters 1.2\%) achieves 88.01\% val.\ acc., but is not at a 6\% or 12\% budget. A trajectory-only (1.2\%) ablation (frozen or minimal instruction encoder) and strict matched-budget comparisons (e.g.\ 1--2\% or 6--12\% across methods) are left for future work.

\subsection{Visualizations}

\begin{figure*}[h]
\centering
\includegraphics[width=0.9\textwidth]{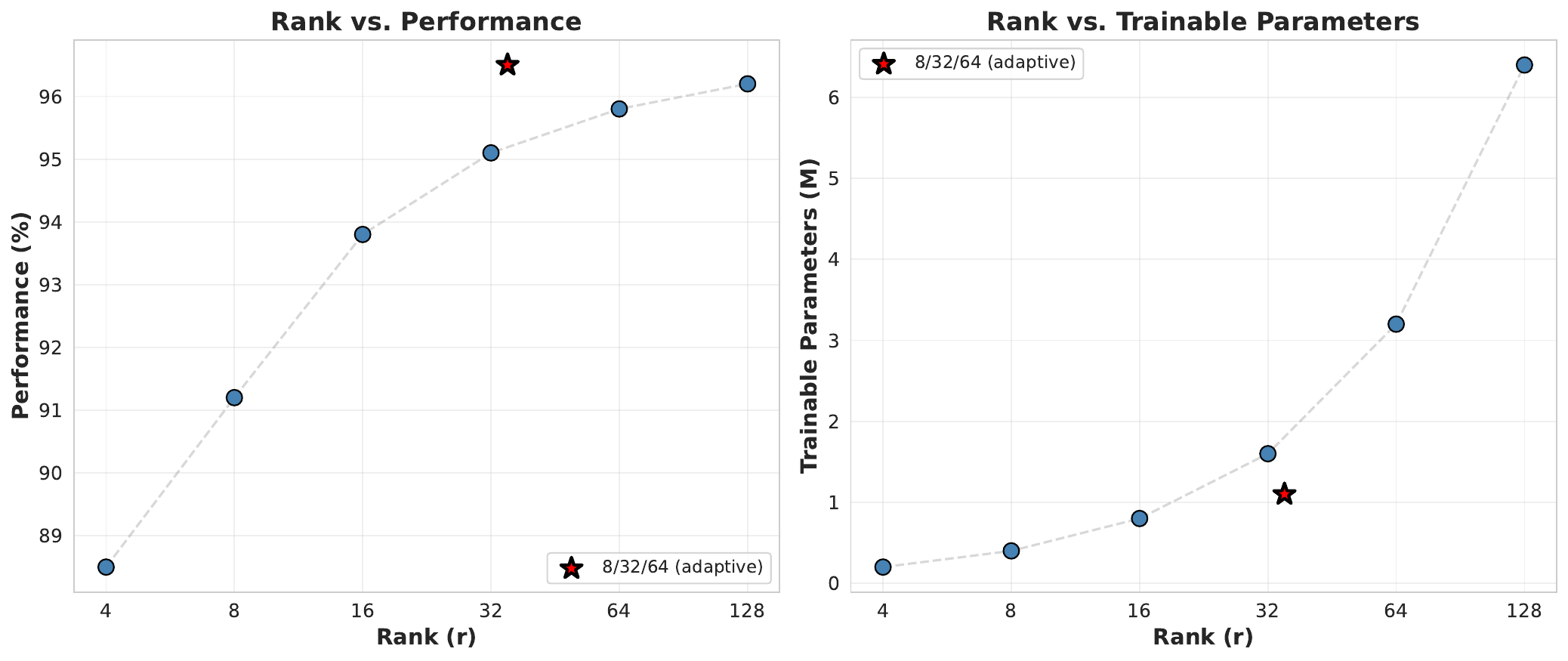}
\caption{Rank vs.\ performance (left) and vs.\ trainable parameters (right). Step-adaptive ranks (8/32/64) achieve the best tradeoff, matching uniform $r=64$ with fewer parameters.}
\label{fig:rank_ablation}
\end{figure*}

\begin{figure*}[h]
\centering
\includegraphics[width=0.7\textwidth]{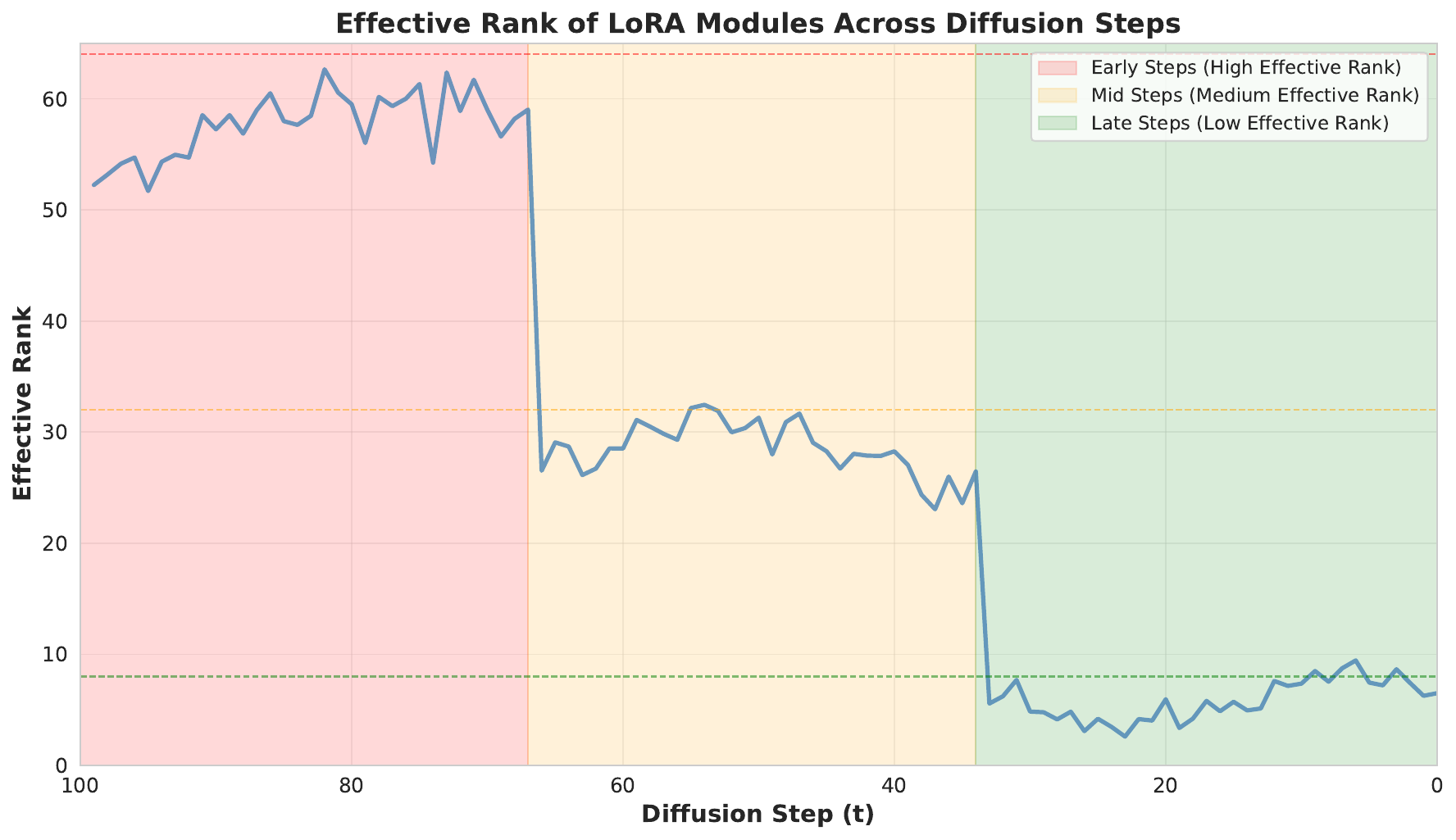}
\caption{Effective rank of LoRA modules across diffusion steps. Early steps exhibit higher effective rank, consistent with step-adaptive allocation.}
\label{fig:effective_rank}
\end{figure*}

Figure~\ref{fig:rank_ablation} plots performance and trainable parameters versus rank configuration, comparing step-adaptive ranks with uniform settings. Figure~\ref{fig:effective_rank} shows the effective rank of LoRA modules across diffusion steps, validating our step-adaptive allocation strategy. Section~\ref{sec:data_efficiency} reports data efficiency (Figure~\ref{fig:data_efficiency}, Table~\ref{tab:data_efficiency}). 

We computed effective rank, defined as the entropy of normalized singular values, for early, middle, and late denoising phases. Despite FiLM conditioning, the empirical effective rank remains bounded by the nominal bottleneck dimension $r$. Early denoising steps exhibit higher effective rank than later steps, which supports the use of step-adaptive rank allocation (Figure~\ref{fig:effective_rank}).

\subsection{Data Efficiency}
\label{sec:data_efficiency}

\begin{figure*}[h]
\centering
\includegraphics[width=0.7\textwidth]{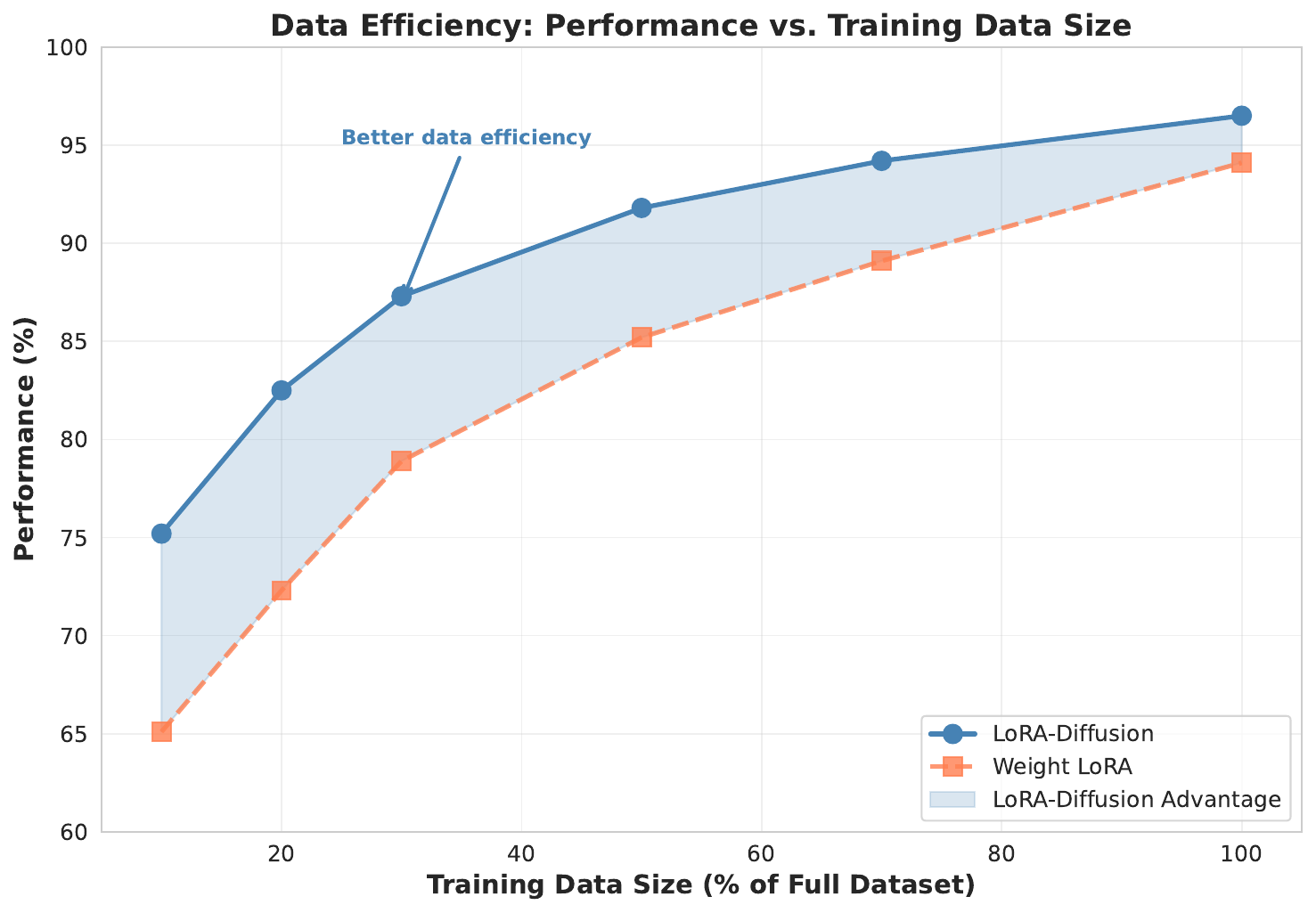}
\caption{Performance vs.\ training data size (SST-2 validation accuracy). LoRA-Diffusion and weight LoRA trained at 10\%, 20\%, 40\%, 60\%, 80\%, and 100\% of the training set. Results from job 44079308 (seed 42). Both methods plateau at 20\% data, indicating efficient convergence with limited training samples.}
\label{fig:data_efficiency}
\end{figure*}

We train LoRA-Diffusion and weight LoRA on SST-2 at 10\%, 20\%, 40\%, 60\%, 80\%, and 100\% of the training set (10k steps per run, seed 42). Results from job 44079308 are shown in Figure~\ref{fig:data_efficiency} and Table~\ref{tab:data_efficiency}. LoRA-Diffusion reaches 90.3\% token-level val accuracy with only 10\% of the data and plateaus near 91.2\% from 20\% onward; weight LoRA plateaus near 83.9\% from 20\% onward. The identical results from 20\% to 100\% indicate that both methods converge to their validation accuracy plateau with approximately 20\% of the training data (13,470 samples), demonstrating efficient learning where additional data beyond this point does not improve performance. LoRA-Diffusion thus achieves higher accuracy at every data fraction and is more data-efficient, particularly in the low-data regime (10\%). This suggests that trajectory-level adaptation can leverage limited supervision more effectively than weight-level LoRA for this diffusion setup.

\begin{table}[ht]
\centering
\caption{SST-2 Data Efficiency}
\label{tab:data_efficiency}
\begin{tabular}{@{}lcccc@{}}
\toprule
\textbf{Data (\%)} & \multicolumn{2}{c}{LoRA-Diffusion} & \multicolumn{2}{c}{Weight LoRA} \\
\cmidrule(lr){2-3} \cmidrule(lr){4-5}
 & Token & Class head & Token & Class head \\
\midrule
10  & 90.3 & 90.3 & 84.1 & 84.1 \\
20  & 91.2 & 91.2 & 83.9 & 83.9 \\
40  & 91.2 & 91.2 & 83.9 & 83.9 \\
60  & 91.2 & 91.2 & 83.9 & 83.9 \\
80  & 91.2 & 91.2 & 83.9 & 83.9 \\
100 & 91.2 & 91.2 & 83.9 & 83.9 \\
\bottomrule
\end{tabular}
\end{table}


\section{Conclusion}
\label{sec:conclusion}

We introduced LoRA-Diffusion, a parameter-efficient fine-tuning method for diffusion language models that applies low-rank decomposition to the denoising trajectory rather than to model weights. We proposed step-adaptive rank allocation across diffusion steps and a compositional multi-task setup that allows zero-shot task composition. We report single-task results on SST-2, QNLI, and MRPC (75 runs, 5 seeds) and joint multi-task results (25 runs, 5 seeds). On SST-2, LoRA-Diffusion achieves the highest mean token-level denoising validation accuracy (88.01\%), indicating more effective learning of task-specific diffusion trajectories. Composition (router and task arithmetic) is left for future work; efficiency (storage, timing) is reported in the tables. We provide ablations for rank and orthogonality regularization. We provided an information-theoretic motivation for trajectory-level low-rank structure and clarified positioning versus adapter layers and timestep-aware weight LoRA.

\textbf{Limitations:} Our evaluation covers three GLUE tasks (SST-2, QNLI, MRPC) and joint multi-task training at a single model size (137.7M parameters). We do not compare at strictly matched parameter budgets (e.g.\ 1--2\% or 6--12\%); isolating the contribution of the instruction encoder (27.5\%) versus the trajectory adapters (1.2\%) would require ablations with a frozen or minimal encoder. Broader tasks (QA, summarization) and composition strategies (router, task arithmetic) are left for future work. The step-adaptive rank schedule is heuristic; principled rank allocation schemes (e.g., GeLoRA-style Fisher-based ranks) could be integrated. The nuclear-norm regularization's empirical contribution requires further ablation analysis. Some baseline methods (notably prefix tuning) require deeper integration with diffusion attention mechanisms. Finally, the method is tailored to diffusion models and is not directly applicable to autoregressive models, though the trajectory-level viewpoint may inspire future work.

Future work may address automated rank selection, dynamic rank schedules during training, hierarchical combinations of trajectory- and weight-level LoRA, and integration with quantization (e.g. QLoRA-style). Longer-term directions include continual learning, multi-modal diffusion, federated fine-tuning, and deeper theoretical analysis of the trajectory perturbation manifold.

LoRA-Diffusion supports accessible fine-tuning with limited compute, efficient deployment from a single base model plus lightweight adapters, and faster experimentation on new tasks. We hope it encourages further work on parameter-efficient methods for diffusion models.

Code, configurations, and evaluation scripts are available at\\
\url{https://github.com/ikhazra/lora-diffusion}.
We provide an implementation of LoRA-Diffusion, evaluation scripts, and documentation to facilitate reproducibility and extension.

\subsection*{Reproducibility}
We use PyTorch 2.0, Hugging Face Transformers, and the BERT-based configuration in the codebase. Base model: 137.7M trainable parameters (12 layers, 768 hidden, 12 heads). Diffusion: $T=100$ steps, cosine schedule. LoRA-Diffusion: $\lambda_{\text{rank}}=0.01$, $\lambda_{\text{orth}}=0.001$, lr $10^{-4}$, batch 64. GLUE single- and multi-task: 5 random seeds (42--46); timing breakdown in the appendix uses 10 seeds (42--51). Scripts accept \texttt{--seed} and \texttt{--num-seeds}. Data: SST-2, QNLI, MRPC from Hugging Face datasets. Hardware: 4$\times$A100 40GB. Code and configs: \url{https://github.com/ikhazra/lora-diffusion}.

\bibliographystyle{IEEEtran}
\bibliography{reference}

@article{aghajanyan2020intrinsic,
  author    = {Aghajanyan, Armen and Zettlemoyer, Luke and Gupta, Sonal},
  title     = {Intrinsic dimensionality explains the effectiveness of language model fine-tuning},
  journal   = {arXiv preprint arXiv:2012.13255},
  year      = {2020}
}

@inproceedings{austin2021structured,
  author    = {Austin, Jacob and Johnson, Daniel D. and Ho, Jonathan and Tarlow, Daniel and {Van Den Berg}, Rian},
  title     = {Structured denoising diffusion models in discrete state-spaces},
  booktitle = {Advances in Neural Information Processing Systems},
  volume    = {34},
  pages     = {17981--17993},
  year      = {2021}
}

@inproceedings{brown2020language,
  author    = {Brown, Tom and Mann, Benjamin and Ryder, Nick and Subbiah, Melanie and Kaplan, Jared D. and Dhariwal, Prafulla and Neelakantan, Arvind and Shyam, Pranav and Sastry, Girish and Askell, Amanda},
  title     = {Language models are few-shot learners},
  booktitle = {Advances in Neural Information Processing Systems},
  volume    = {33},
  pages     = {1877--1901},
  year      = {2020}
}

@article{dettmers2023qlora,
  author    = {Dettmers, Tim and Pagnoni, Artidoro and Holtzman, Ari and Zettlemoyer, Luke},
  title     = {{QLoRA}: Efficient finetuning of quantized {LLMs}},
  journal   = {arXiv preprint arXiv:2305.14314},
  year      = {2023}
}

@article{fedus2022switch,
  author    = {Fedus, William and Zoph, Barret and Shazeer, Noam},
  title     = {Switch transformers: Scaling to trillion parameter models with simple and efficient sparsity},
  journal   = {Journal of Machine Learning Research},
  volume    = {23},
  number    = {120},
  pages     = {1--39},
  year      = {2022}
}

@inproceedings{hoogeboom2021autoregressive,
  author    = {Hoogeboom, Emiel and Nielsen, Didrik and Jaini, Priyank and Forr{\'e}, Patrick and Welling, Max},
  title     = {Argmax flows and multinomial diffusion: Learning categorical distributions},
  booktitle = {Advances in Neural Information Processing Systems},
  volume    = {34},
  pages     = {12454--12465},
  year      = {2021}
}

@inproceedings{houlsby2019parameter,
  author    = {Houlsby, Neil and Giurgiu, Andrei and Jastrz{\k{e}}bski, Stanis{\l}aw and Morrone, Bruna and {De Laroussilhe}, Quentin and Gesmundo, Andrea and Attariyan, Mona and Gelly, Sylvain},
  title     = {Parameter-efficient transfer learning for {NLP}},
  booktitle = {International Conference on Machine Learning},
  pages     = {2790--2799},
  year      = {2019}
}

@article{hu2021lora,
  author    = {Hu, Edward J. and Shen, Yelong and Wallis, Phillip and Allen-Zhu, Zeyuan and Li, Yuanzhi and Wang, Shean and Wang, Lu and Chen, Weizhu},
  title     = {{LoRA}: Low-rank adaptation of large language models},
  journal   = {arXiv preprint arXiv:2106.09685},
  year      = {2021}
}

@article{ilharco2022editing,
  author    = {Ilharco, Gabriel and Ribeiro, Marco Tulio and Wortsman, Mitchell and Gururangan, Suchin and Schmidt, Ludwig and Hajishirzi, Hannaneh and Farhadi, Ali},
  title     = {Editing models with task arithmetic},
  journal   = {arXiv preprint arXiv:2212.04089},
  year      = {2022}
}

@inproceedings{lester2021power,
  author    = {Lester, Brian and Al-Rfou, Rami and Constant, Noah},
  title     = {The power of scale for parameter-efficient prompt tuning},
  booktitle = {Proceedings of the 2021 Conference on Empirical Methods in Natural Language Processing},
  pages     = {3045--3059},
  year      = {2021}
}

@inproceedings{li2018measuring,
  author    = {Li, Chunyuan and Farkhoor, Heerad and Liu, Rosanne and Yosinski, Jason},
  title     = {Measuring the intrinsic dimension of objective landscapes},
  booktitle = {International Conference on Learning Representations},
  year      = {2018}
}

@inproceedings{li2021prefix,
  author    = {Li, Xiang Lisa and Liang, Percy},
  title     = {Prefix-tuning: Optimizing continuous prompts for generation},
  booktitle = {Proceedings of the 59th Annual Meeting of the Association for Computational Linguistics},
  pages     = {4582--4597},
  year      = {2021}
}

@inproceedings{li2022diffusion,
  author    = {Li, Xiang Lisa and Thickstun, John and Gulrajani, Ishaan and Liang, Percy S. and Hashimoto, Tatsunori B.},
  title     = {Diffusion-{LM} improves controllable text generation},
  booktitle = {Advances in Neural Information Processing Systems},
  volume    = {35},
  pages     = {4328--4343},
  year      = {2022}
}

@inproceedings{lou2023discrete,
  author    = {Lou, Aaron and Meng, Chenlin and Ermon, Stefano},
  title     = {Discrete diffusion modeling by estimating the ratios of the data distribution},
  booktitle = {International Conference on Machine Learning},
  pages     = {22481--22505},
  year      = {2023}
}

@article{sahoo2024masked,
  author    = {Sahoo, Prateek and Nguyen, Hieu and Loh, Chien-Yu and Kumar, Ashish and Narasimhan, Karthik},
  title     = {Simple and effective masked diffusion language models},
  journal   = {arXiv preprint arXiv:2406.07524},
  year      = {2024}
}

@article{tishby2000information,
  author    = {Tishby, Naftali and Pereira, Fernando C. and Bialek, William},
  title     = {The information bottleneck method},
  journal   = {arXiv preprint physics/0004057},
  year      = {2000}
}

@inproceedings{tishby2015deep,
  author    = {Tishby, Naftali and Zaslavsky, Noga},
  title     = {Deep learning and the information bottleneck principle},
  booktitle = {IEEE Information Theory Workshop},
  pages     = {1--5},
  year      = {2015}
}

@article{wang2020orthogonal,
  author    = {Wang, Zifan and Zhang, Zichen and Lee, Chen-Yu and Zhang, Han and Sun, Ruixin and Ren, Xiang and Su, Guande and Perot, Vincent and Dy, Jennifer and Pfister, Tomas},
  title     = {Learning to prompt for continual learning},
  journal   = {arXiv preprint arXiv:2112.08654},
  year      = {2020}
}

@article{zaken2021bitfit,
  author    = {Zaken, Elad Ben and Ravfogel, Shauli and Goldberg, Yoav},
  title     = {{BitFit}: Simple parameter-efficient fine-tuning for transformer-based masked language-models},
  journal   = {arXiv preprint arXiv:2106.10199},
  year      = {2021}
}

@inproceedings{zhang2023adalora,
  author    = {Zhang, Qian and Chen, Muxin and Bukharin, Alexander and He, Pengcheng and Cheng, Yu and Chen, Weizhu and Zhao, Tuo},
  title     = {{AdaLoRA}: Adaptive budget allocation for parameter-efficient fine-tuning},
  booktitle = {International Conference on Learning Representations},
  year      = {2023}
}

@inproceedings{foura2024,
  title     = {FouRA: Fourier Low Rank Adaptation},
  author    = {Borse, Shreyas and Riti, Pierfrancesco and Bhalodia, Riddhish and Porikli, Fatih},
  booktitle = {Advances in Neural Information Processing Systems},
  year      = {2024}
}

@inproceedings{efficientdm2023,
  title     = {EfficientDM: Efficient Quantization-Aware Fine-Tuning of Low-Bit Diffusion Models},
  author    = {He, Yefei and Liu, Jing and Wu, Weijia and Zhou, Hong and Zhuang, Bohan},
  booktitle = {International Conference on Learning Representations},
  year      = {2024}
}

@misc{selora2024,
  title         = {SeLoRA: Self-Expanding Low-Rank Adaptation of Latent Diffusion Model for Medical Image Generation},
  author        = {Mao, Yuchen},
  year          = {2024},
  eprint        = {2408.07196},
  archivePrefix = {arXiv},
  primaryClass  = {cs.CV}
}

@article{gelora2024,
  title         = {GeLoRA: Geometric Adaptive Ranks For Efficient LoRA Fine-tuning},
  author        = {Ed-dib, Abdessalam and Datbayev, Zhanibek and Aboussalah, Amine Mohamed},
  journal       = {arXiv preprint arXiv:2412.09250},
  year          = {2024},
  eprint        = {2412.09250},
  archivePrefix = {arXiv},
  primaryClass  = {cs.LG}
}

@article{tclora2024,
  title         = {TC-LoRA: Temporally Modulated Conditional LoRA for Adaptive Diffusion Control},
  author        = {Cho, Minkyoung and Ohana, Ruben and Jacobsen, Christian and Jothi, Adityan and Chen, Min-Hung and Mao, Z. Morley and Can, Ethem},
  journal       = {arXiv preprint arXiv:2510.09561},
  year          = {2025},
  eprint        = {2510.09561},
  archivePrefix = {arXiv},
  primaryClass  = {cs.CV}
}

@inproceedings{zhao2025msfp,
  title     = {Pioneering 4-Bit FP Quantization for Diffusion Models: Mixup-Sign Quantization and Timestep-Aware Fine-Tuning},
  author    = {Zhao, Maosen and Chen, Pengtao and Yu, Chong and Wen, Yan and Tan, Xudong and Chen, Tao},
  booktitle = {Proceedings of the IEEE/CVF Conference on Computer Vision and Pattern Recognition},
  year      = {2025}
}

@article{dong2025glance,
  title={Glance: Accelerating Diffusion Models with 1 Sample},
  author={Dong, Zhuobai and Zhao, Rui and Wu, Songjie and Yi, Junchao and Li, Linjie and Yang, Zhengyuan and Wang, Lijuan and Wang, Alex Jinpeng},
  journal={arXiv preprint arXiv:2512.02899},
  year={2025}
}

@article{gao2025delta,
  title={Delta Sampling: Data-Free Knowledge Transfer Across Diffusion Models},
  author={Gao, Zhidong and Pan, Zimeng and Yao, Yuhang and Xie, Chenyue and Wei, Wei},
  journal={arXiv preprint arXiv:2512.03056},
  year={2025}
}

@article{zhang2025subject,
  title={Subject or Style: Adaptive and Training-Free Mixture of LoRAs},
  author={Zhang, Jia-Chen and Xiong, Yu-Jie},
  journal={arXiv preprint arXiv:2508.02165},
  year={2025}
}

@inproceedings{khazrak2025addressing,
  title={Addressing Small and Imbalanced Medical Image Datasets Using Generative Models},
  author={Khazrak, Iman and Takhirova, Shakhnoza and Rezaee, Mostafa M and Yadollahi, Mehrdad and Green II, Robert C and Niu, Shuteng},
  booktitle={Artificial Intelligence and Applications},
  year={2025}
}

@article{khazrak2025feasibility,
  title={Feasibility of improving vocal fold pathology image classification with synthetic images generated by DDPM-based GenAI: a pilot study},
  author={Khazrak, Iman and Zainaee, Shahryar and M. Rezaee, Mostafa and Ghasemi, Mehran and C. Green, Robert},
  journal={European Archives of Oto-Rhino-Laryngology},
  pages={1--15},
  year={2025},
  publisher={Springer}
}

@inproceedings{soboleva2026t,
  title={T-lora: Single image diffusion model customization without overfitting},
  author={Soboleva, Vera and Alanov, Aibek and Kuznetsov, Andrey and Sobolev, Konstantin},
  booktitle={Proceedings of the AAAI Conference on Artificial Intelligence},
  volume={40},
  number={11},
  pages={9051--9059},
  year={2026}
}

\end{document}